\documentclass[letterpaper, 10 pt, conference]{ieeeconf}
\IEEEoverridecommandlockouts    
\usepackage{multirow}
\usepackage{booktabs}
\usepackage{makecell}
\usepackage{flushend}

\usepackage{graphics}           
\usepackage{times}              
\usepackage{amsmath}            
\usepackage{amssymb}            
\usepackage{graphicx}
\usepackage{algorithm}
\usepackage[noend]{algpseudocode}
\usepackage{booktabs}
\usepackage{color}
\usepackage{listings}
\usepackage{subfiles}
\usepackage{hyperref}
\usepackage[nocompress]{cite} 
\definecolor{instructioncolor}{rgb}{.5,.5,.5}

\usepackage[font=small]{caption}

\def\figref#1{Fig.~\ref{#1}}
\def\tabref#1{Tab.~\ref{#1}}
\def\eqref#1{Eq.~(\ref{#1})}

\makeatletter
\usepackage{xspace}
\DeclareRobustCommand\onedot{\futurelet\@let@token\@onedot}
\def\@onedot{\ifx\@let@token.\else.\null\fi\xspace}

\def\etal{{et al}\onedot}
\makeatother

\def\etalcite#1{\etal~\cite{#1}}

\usepackage{array}
\newcolumntype{L}[1]{>{\raggedright\let\newline\\\arraybackslash\hspace{0pt}}m{#1}}
\newcolumntype{C}[1]{>{\centering\let\newline\\\arraybackslash\hspace{0pt}}m{#1}}
\newcolumntype{R}[1]{>{\raggedleft\let\newline\\\arraybackslash\hspace{0pt}}m{#1}}

\def\argmin{\mathop{\rm argmin}}

\newcommand{\RR}{\mathbb{R}}

\renewcommand{\b}[1]{\mbox{\boldmath$#1$}}

\renewcommand{\v}[1]{{\b #1}} 

\newcommand{\m}[1]{{\mbox{{\sffamily\slshape{#1\/}}}}}

\makeatletter
\def\blfootnote{\gdef\@thefnmark{}\@footnotetext}
\makeatother

\title{\fontsize{16pt}{18pt}\selectfont\bfseries Loop Closure via Maximal Cliques in 3D LiDAR-Based SLAM}

\author{Javier Laserna$^{1}$ \and Saurabh Gupta$^{2}$ \and Oscar Martinez Mozos$^{1}$ \and Cyrill Stachniss$^{2}$ \and Pablo San Segundo$^{1}$%
}

\begin{document}
\thispagestyle{empty}
\pagestyle{empty}

\twocolumn[{
	\renewcommand\twocolumn[1][]{#1}
	\maketitle
	\begin{center}
		\includegraphics[width=0.9\textwidth]{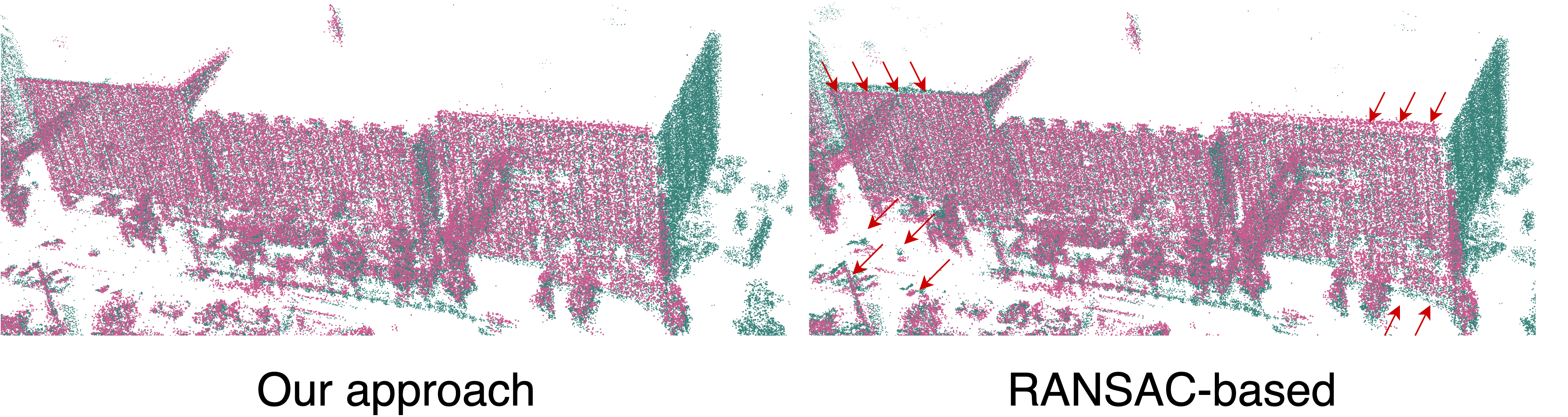}
    	\captionof{figure}{Successful loop closure association between two local LiDAR submaps in a complex urban environment. The red and green points represent keypoints from the query and reference maps, respectively. The left image shows the result of our approach, while the right image shows the result of a RANSAC-based method.}\label{fig:local_map_association}
	\end{center}%
	\vspace{2.5mm}
}]

\begin{abstract}
	Reliable loop closure detection remains a critical challenge in 3D LiDAR-based SLAM, especially under sensor noise, environmental ambiguity, and viewpoint variation conditions. RANSAC is often used in the context of loop closures for geometric model fitting in the presence of outliers. However, this approach may fail, leading to map inconsistency. We introduce a novel deterministic algorithm, CliReg, for loop closure validation that replaces RANSAC verification with a maximal clique search over a compatibility graph of feature correspondences. This formulation avoids random sampling and increases robustness in the presence of noise and outliers. We integrated our approach into a real-time pipeline employing binary 3D descriptors and a Hamming distance embedding binary search tree-based matching. We evaluated it on multiple real-world datasets featuring diverse LiDAR sensors. The results demonstrate that our proposed technique consistently achieves a lower pose error and more reliable loop closures than RANSAC, especially in sparse or ambiguous conditions. Additional experiments on 2D projection-based maps confirm its generality across spatial domains, making our approach a robust and efficient alternative for loop closure detection.
\end{abstract}

\blfootnote{$^{1}$The authors are with the ETSIDI, Universidad Politécnica de Madrid, 28012 Madrid, Spain.
$^{2}$The authors are with the Center for Robotics, University of Bonn, Germany. Cyrill Stachniss is additionally with the Lamarr Institute for Machine Learning and Artificial Intelligence, Germany.}
\blfootnote{ \textbf{This is the accepted manuscript. The final published article is available at \url{https://doi.org/10.1109/ECMR65884.2025.11163179}}}%
\blfootnote{ © 2025 IEEE.  Personal use of this material is permitted.  Permission from IEEE must be obtained for all other uses, in any current or future media, including reprinting/republishing this material for advertising or promotional purposes, creating new collective works, for resale or redistribution to servers or lists, or reuse of any copyrighted component of this work in other works.}

 \section{Introduction}\label{sec:intro}


Loop closure detection is an essential component of simultaneous localization and mapping~(SLAM) systems~\cite{grisetti2010titsmag,behley2018rss,blanco-claraco2025ijrr}, allowing robots and autonomous vehicles to identify locations they have previously visited and correct the global drift in trajectory obtained from sequential odometry sources. A globally consistent trajectory, in return, allows such robots to maintain a consistent map representation of their environment, enabling downstream autonomy tasks such as global re-localization and path planning.

Identifying loop closures can be difficult in large outdoor 3D spaces where LiDAR point clouds are sparse, noisy, and undergo spatio-temporal changes due to dynamic scenes.
Recent state-of-the-art methods for loop closure detection~\cite{yuan2023icra,gupta2024icra,kim2021tro,jiang2023icra-ccas,dube2017icra} rely on computing local and/or global feature descriptors from LiDAR point clouds, narrowing the search space of loop closure candidates through a feature matching process. Subsequently, they apply geometric verification of these candidate loop closures to find consistent 3D pose constraints for the pose-graph optimization in SLAM\@. Such geometric verification can be challenging due to outlier correspondences, especially in purely feature-based correspondence sets generated by typical loop closure methods~\cite{gupta2024icra,luo2021ral,dube2017icra}, leading to incorrect loop closures. Already, a few wrong closures can substantially affect the overall performance of the SLAM system.

Among the various strategies for geometric verification of loop closure candidates, outlier rejection techniques like RANSAC~\cite{fischler1981cacm} are widely used, as they iteratively sample minimal sets of correspondences to fit a transformation model while handling outliers. However, this approach can be sensitive to outlier correspondences and may fail to find the correct loop closure in challenging environments. Moreover, such a sampling-based approach can be computationally expensive in the presence of a large number of outlier correspondences.

The main contribution of this paper is a loop closure validation based on a strong combinatorial optimization backend. We propose replacing the RANSAC-based verification process with a maximal clique search algorithm CliReg~\cite{laserna2025tro} to verify candidate loop closures. Our approach aims to find the largest mutually consistent set of feature correspondences with a graph-theoretic approach to achieve geometric consistency without relying on random sampling. Our method provides a principled alternative to traditional consensus-based verification with improved robustness and precision.
As shown in~\figref{fig:local_map_association}, our approach produces a consistent set of feature matches between local maps. The resulting geometric transformation is directly usable as a loop closure constraint in pose-graph optimization. We propose a complete loop closure detection and validation pipeline using binary 3D descriptors and an efficient binary search tree-based database for appearance matching. We evaluate the method on challenging 3D LiDAR datasets and show that it achieves better pose accuracy than RANSAC-based geometric verification while being computationally efficient for real-time SLAM\@. We demonstrate that our approach successfully validates loop closures even when RANSAC fails to find consistent inlier correspondences. Our method improves alignment accuracy and robustness without introducing runtime overheads that hinder real-time use. These characteristics make it a compelling alternative for loop closure validation in LiDAR-based SLAM systems.

 \section{Related Work}\label{sec:related}

 The precision and robustness of loop closure detection within any SLAM pipeline directly influence mapping quality and localization accuracy. Traditional loop closure detection methods for LiDAR SLAM rely on geometric feature-based algorithms and corresponding descriptors such as SHOT~\cite{salti2014cviu}, FPFH~\cite{rusu2009icra}, NARF~\cite{steder2010irosws}, as well as point feature-based place recognition~\cite{steder2010icra}. Although effective in controlled environments, these techniques often suffer from reduced recall under viewpoint changes, sensor noise, or environmental variability. More recent pipelines leverage global place recognition or voting schemes~\cite{bosse2013icra,dube2017icra}, but they typically depend on robust feature filtering for precise loop closure validation. Efforts to improve robustness across diverse sensor setups have also led to new benchmarks such as HeLiPR~\cite{jung2024ijrr} which support place recognition evaluation in heterogeneous LiDAR configurations.

 Robust loop closure detection plays a central role in maintaining global map consistency and reducing drift in SLAM systems, especially under large-scale 3D LiDAR-based scenarios. The literature proposes a wide range of approaches, from geometric descriptor matching to learned place recognition, each with distinct robustness, efficiency, and generalization trade-offs.
 
 RANSAC-based techniques~\cite{fischler1981cacm,yang2015pami} remain the most widely used solution for outlier rejection within geometric validation of candidate loop closures. Despite their popularity, they rely on random sampling and often fail under high outlier ratios or sparse correspondence sets~\cite{raguram2008eccv}. Alternative robust verification methods such as pairwise consistent measurement have also been explored in multi-robot SLAM~\cite{mangelson2018icra}. This motivates the use of deterministic alternatives. Earlier work on deterministic methods includes branch-and-bound strategies for data association, such as the joint compatibility branch and bound algorithm by Neira and Tardós~\cite{neira2001tro}. Olson~\etalcite{olson2009icra} proposed a real-time correlative scan matching strategy, influencing early SLAM pipelines.
 
 Clique-based methods have recently been introduced to generate robust data associations for point cloud registration as an alternative to sampling-based methods. In particular, maximal clique search can extract mutually consistent correspondences under rigid-body constraints. Recent research has focused on combinatorial optimization and graph-based methods~\cite{shi2021icra-raga}, particularly maximal clique enumeration techniques. While earlier works such as~\cite{segundo2015cor,segundo2016cor,segundo2019cor,segundo2016informatica} studied the theoretical aspects of clique enumeration and introduced graph pruning strategies for general-purpose applications, their focus was not on 3D registration. The CliReg algorithm~\cite{laserna2025tro} applies maximal clique search specifically for 3D point cloud registration, demonstrating significant gains in robustness and accuracy.
 
 Efficient indexing and retrieval of binary descriptors are also crucial to the clique-based correspondence graph generation methods. HBST~\cite{schlegel2018ral} enables scalable feature lookup using binary trees, and its use with binary descriptors like ORB~\cite{rublee2011iccv} and B-SHOT~\cite{prakhya2015iros} has shown promise in place recognition.
 
 Learning-based solutions~\cite{angelina2018cvpr,xu2022cis} offer high retrieval recall but suffer from training dependency and limited generalization. While these models can perform well on curated datasets, their robustness in diverse and previously unseen real-world conditions remains a challenge~\cite{lowry2016tro}. Earlier works explored binary vocabularies~\cite{galvez2012tro} and nonlinear embedding for nearest neighbor classification~\cite{salakhutdinov2007icais}.
 
 In summary, the state of the art reveals a gap between robustness and efficiency. Our approach bridges this by integrating binary 3D descriptors with deterministic clique-based filtering. Unlike prior methods, our approach looks for mutual consistency among feature matches, reducing false positives and improving loop closure reliability without sacrificing runtime performance.

\section{CliReg-Based Loop Closures}\label{sec:main}

\begin{figure*}[t]
	\centering
	\includegraphics[width=0.9\textwidth]{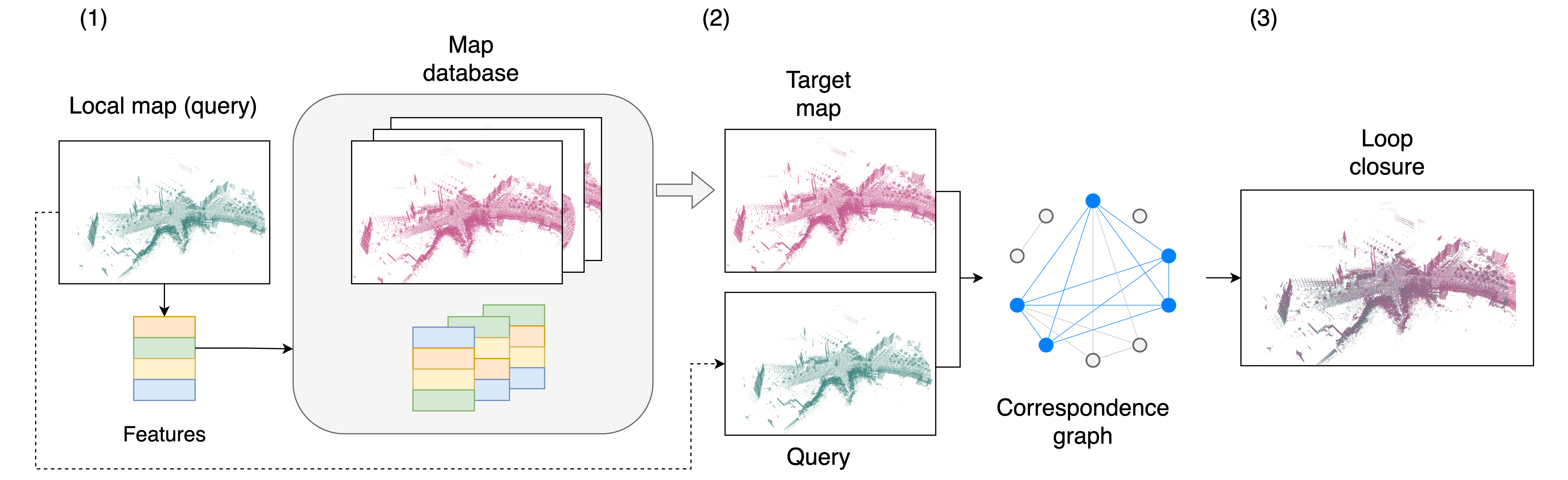}
	\caption{An overview of our loop closure detection and validation pipeline. It consists of three stages: (1) Feature Extraction and Encoding, (2) Correspondence Graph generation, and (3) Pose Estimation via maximal clique search. This integration enables efficient and robust loop closure detection across 2D and 3D feature representations.}\label{fig:overview}
	\vspace{-10pt}
\end{figure*}

We present a loop closure validation pipeline that replaces RANSAC with a clique-based approach for geometric verification. As illustrated in Fig.~\ref{fig:overview}, our pipeline comprises three main stages: (1)~feature extraction and encoding using binary descriptors, (2)~construction of a correspondence graph based on descriptor matches, and (3)~pose estimation through maximal clique search and least-squares alignment. Our approach identifies the largest subset of geometrically consistent correspondences without relying on random sampling.

\subsection{Preliminaries}

Let~$\mathcal{M}$ and~$\mathcal{Q}$ denote two local maps composed of 3D point clouds or their 2D projections. From both maps, we extract a set of feature keypoints along with their associated descriptors. Based on descriptor matching, we then establish a tentative set of point-to-point correspondences~$\mathcal{C} = {\v{m}_i, \v{q}_i}$, where~$\v{m}_i \in \mathcal{M}$ and~$\v{q}_i \in \mathcal{Q}$.

Assuming a rigid body transformation model, we expect inlier correspondences to satisfy the following relation:

\begin{equation}\label{mod:rigid}
	\v{q}_i = \m{R} \v{m}_i + \v{t} + \v{\epsilon}_i, \quad i \in \{1, \dots, |\mathcal{C}|\},
\end{equation}

\noindent where~$\m{R} \in \mathbb{SO}(n)$ is a rotation matrix,~$\v{t} \in \RR^n$ is a translation vector,~$\v{\epsilon}_i \in \RR^n$ is an additive noise term, and~$n \in \{2, 3\}$ denotes the spatial dimension~(depending on whether 3D point clouds or 2D image projections are used). 

The validation of matches aims to find the largest subset of correspondences consistent with a single rigid body transformation.

\subsection{Clique-Based Registration}

We construct a correspondence graph where each node represents a candidate correspondence, and edges connect mutually consistent pairs~(i.e., pairs satisfying the rigidity constraint in~\eqref{mod:rigid} within some tolerance). We then employ a branch-and-bound search to identify the maximum clique of mutually consistent correspondences, denoted by~$\mathcal{C}^* \subseteq \mathcal{C}$.

Given the inlier set~$\mathcal{C}^*$, we estimate the optimal transformation parameters~$(\m{R}^*, \v{t}^*)$ by solving the following least-squares problem:

\begin{equation}\label{opt:rigid}
	\m{R}^*,\ \v{t}^* = \argmin_{\m{R},\ \v{t}} \sum_{(\v{m}_i, \v{q}_i) \in \mathcal{C}^*} ||\v{q}_i - \m{R}\v{m}_i - \v{t}||^2.
\end{equation}

Researchers have extensively documented a closed-form solution for~\eqref{opt:rigid} in the literature~\cite{arun1987pami,horn1987josa,walker1991cvgip}. If~$|\mathcal{C}^*|$ exceeds a predefined inlier threshold, we accept the estimated transformation as a valid loop closure and directly integrate it as a constraint in pose graph optimization.

Unlike sampling-based methods such as RANSAC, CliReg provides a deterministic and globally optimal inlier set under the rigid body transformation, making it particularly robust in scenarios with high outlier rates.

\subsection{Feature Extraction and Encoding}

For each local map, we extract keypoints and compute descriptors depending on the dimensionality. In 3D, we detect keypoints using intrinsic shape signatures~(ISS) algorithm~\cite{zhong2009iccvws} directly on voxelized LiDAR point clouds. We describe them using SHOT descriptors~\cite{salti2014cviu}, and then apply a median thresholding strategy to binarize them into compact B-SHOT descriptors~\cite{prakhya2015iros}.

In the 2D setting, we follow Gupta~\etalcite{gupta2024icra}, first projecting the 3D LiDAR point clouds into bird's eye view~(BEV) density images and extracting ORB descriptors~\cite{rublee2011iccv}, which provide binary descriptors suitable for fast matching. Regardless of the representation, we organize all binary descriptors using an HBST data structure~\cite{schlegel2018ral} to enable real-time insertion and nearest-neighbor lookup via Hamming distance metric.

\subsection{Descriptor Database and Correspondence Graph Construction}

From the set of tentative matches~$\mathcal{C} = \{\v{m}_i, \v{q}_i\}$, we build a correspondence graph~$G = (V, E)$ where each vertex represents a descriptor match between $\mathcal{M}$ and $\mathcal{Q}$. An edge connects two nodes if their associated correspondences are mutually compatible, meaning the rigid transformation preserves their pairwise distances. Formally, two correspondences $(\v{m}_i, \v{q}_i)$ and $(\v{m}_j, \v{q}_j)$, we consider them consistent if the following condition is met:

\begin{equation}
    \big| \|\v{m}_i - \v{m}_j\| - \|\v{q}_i - \v{q}_j\| \big| < \epsilon. \label{eq:consistency}
\end{equation}

This constraint preserves pairwise distances under a rigid-body model, enforcing geometric consistency. The threshold~$\epsilon$ is set relative to the voxel resolution. A clique thus represents mutually consistent correspondences explainable by a single transformation.

\begin{figure}[t]
	\centering
	\includegraphics[width=0.24\textwidth]{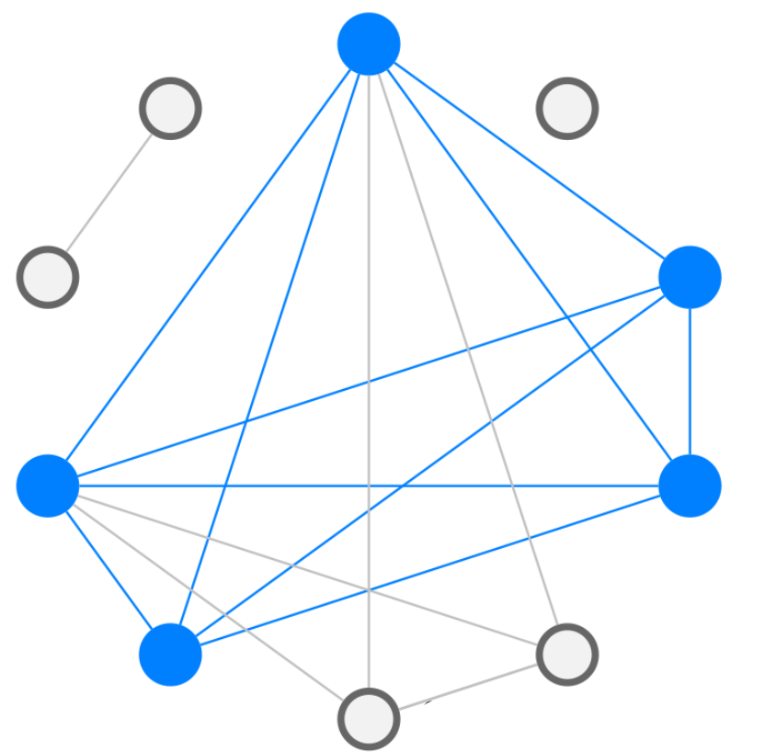}
	\caption{Example of a correspondence graph~$G$ with nodes representing feature matches and edges denoting geometric consistency. The highlighted 5-clique forms the basis for~$\mathbb{SE}(3)$ transformation.}\label{fig:correspondence_graph}
	\vspace{-10pt}
\end{figure}

\figref{fig:correspondence_graph} shows an example of such a correspondence graph, where the blue subgraph represents a maximal clique selected to estimate the rigid body transformation.

\section{Experimental Evaluation}
\label{sec:exp}

We evaluated the effectiveness of our CliReg-based loop closure validation pipeline in the Bridge01, Bridge02 and Roundabout01 sequences of the HeLiPR dataset~\cite{jung2024ijrr}, which include realistic loop closures in urban semi-structured environments. We captured these sequences using three different LiDAR sensors: Aeva~(solid-state), Avia~(MEMS), and Ouster~(mechanical), to represent a wide range of sensor characteristics.

Our evaluation focusses on three key aspects: (i) loop closure validation quality (number of geometrically consistent inliers), (ii) accuracy of pose estimation, and (iii) computational efficiency. We compare our CliReg-based verification with a RANSAC-based baseline under identical conditions, using 3D and 2D features.

\subsection{Metrics and Experimental Setup}

To ensure a fair comparison, we configure a standard set of parameters across all experiments. In both the 2D and 3D pipelines, appearance-based descriptor matching uses a Hamming distance threshold of 50\,bits. A loop closure is accepted only if a minimum number of mutually consistent inliers is found: 10 in the 2D case and 5 in the 3D case.

To compute the absolute pose error~(APE) with and without loop closures~(denoted APE w and APE w/o respectively), we follow a consistent evaluation pipeline. We begin by incorporating all detected loop closures into a pose graph optimization using g2o~\cite{kummerle2011icra}, initializing the graph with the trajectory estimated by the KISS-ICP algorithm~\cite{vizzo2023ral}, and the loop closure constraints obtained from our loop closure validation algorithm. After optimization, we compare the corrected poses with the ground-truth trajectory available for each sensor and scene. Finally, we compute the APE using the evo Python toolbox~\cite{grupp2017github}, a widely adopted tool for evaluating odometry and SLAM performance metrics.

\subsection{2D and 3D Pipeline Description}

\textbf{2D pipeline:} We adopt a BEV density map representation of the environment, with each local map discretised into a grid using a resolution of 0.5\,m. We discard low-density regions by applying a threshold of 5\,\% relative to the maximum cell occupancy. Local maps are formed incrementally until the platform travels a fixed distance, of 100\,m, which defines the spatial extent of each local representation. This setup favours compact, viewpoint-agnostic descriptors, enabling efficient matching using HBST\@. A similar representation for loop closure has been proposed by Gupta~\etalcite{gupta2024icra}.

\textbf{3D pipeline:} We employ voxelization with a 0.5\,m resolution to downsample the point cloud, ensuring uniform spatial coverage. Keypoints are extracted using ISS with scale-adaptive parameters: a salient radius of 3\,m, a nonmaximum suppression radius of 2\,m, and an eigenvalue threshold of 0.975. Around these keypoints, SHOT descriptors are computed within a support radius of 2\,m and binarized to obtain B-SHOT signatures. Geometric consistency is enforced using a pairwise distance between the aligned features with tolerance set to 1\,m. This configuration balances descriptiveness and computational efficiency, making it suitable for real-time SLAM deployments.

\textbf{RANSAC configuration:} For the 3D experiments, we employ RANSAC in which the number of iterations is explicitly fixed to 10,000~(denoted as RANSAC-10K). This decision is motivated by the observation that, when using the number of iterations derived from the classical probabilistic formulation~(assuming an inlier ratio of 0.3 and a success probability of 0.999), RANSAC may fail to detect any loop closures across all evaluated scenes. The same result was observed even when the number of iterations was fixed to 1,000. Therefore, we perform 10,000 RANSAC iterations in all 3D experiments to ensure a meaningful baseline for comparison.

\subsection{Evaluation Results}
\label{sec:exp-3d-results}

\begin{table*}[t] 
\centering
\caption{
\textbf{Loop closure performance using 3D pipeline across sensors and scenes with our method and RANSAC.}
Metrics include number of inliers, runtime, and APE with and without loop closures.
}\label{tab:grouped_by_lidar_full}
\vspace{0.2cm}
\begin{tabular}{llrrrrrr}
\toprule
\textbf{Scene} & \textbf{Sensor} & \textbf{Algorithm} & \textbf{Inliers} & \makecell{\textbf{Mean}\\\textbf{Time (ms)}} & \textbf{APE w (m)} & \textbf{APE w/o (m)} \\
\midrule
\multirow{7}{*}{Bridge01} 
& \multirow{2}{*}{\centering Aeva}  & RANSAC-10K  & -           & -                & -                 & \multirow{2}{*}{\centering 94.67}\\ 
&        & Ours        & \textbf{61} & \textbf{6.25}    & \textbf{27.40}    &\\
\cmidrule(r){2-7}
& \multirow{2}{*}{Avia}   & RANSAC-10K  & 6           & 324.17           & 79.05             & \multirow{2}{*}{\centering 167.83}\\
&        & Ours        & \textbf{24} & \textbf{2.83}    & \textbf{33.98}    &\\
\cmidrule(r){2-7}
& \multirow{2}{*}{Ouster} & RANSAC-10K  & 6           & 317.05           & 50.67             & \multirow{2}{*}{\centering 157.37}\\
&        & Ours        & \textbf{136} & \textbf{4.92}    & \textbf{18.80}   &\\
\specialrule{.1em}{.4em}{.4em}
\multirow{6}{*}{Bridge02}
& \multirow{2}{*}{Aeva}   & RANSAC-10K  & -           & -                & -                 & \multirow{2}{*}{\centering 120.93}\\
&        & Ours        & \textbf{57} & \textbf{4.11}    & \textbf{108.94}   &\\
\cmidrule(r){2-7}
& \multirow{2}{*}{Avia}   & RANSAC-10K  & 21          & 291.31           & \textbf{16.48}    & \multirow{2}{*}{\centering 76.09}\\
&        & Ours        & \textbf{70} & \textbf{3.61}    & 19.00           &  \\
\cmidrule(r){2-7}
& \multirow{2}{*}{Ouster} & RANSAC-10K  & -           & -                & -                 & \multirow{2}{*}{\centering 53.95}\\
&        & Ours        & \textbf{71} & \textbf{4.52}    & \textbf{68.25}    & \\
\bottomrule
\end{tabular}
\end{table*}

We summarize the results for the 3D pipeline in~\tabref{tab:grouped_by_lidar_full}. Our approach consistently yields geometrically more reliable loop closures than RANSAC, particularly in challenging cases where RANSAC fails to detect any closure. Although our approach typically validates fewer matches, the identified inliers exhibit higher mutual consistency and produce a lower absolute pose error~(APE).

In particular, RANSAC often fails entirely in sequences such as Aeva-Bridge01 or Ouster-Bridge02, while our approach reliably detects loop closures with 50–130 inliers. These results illustrate the robustness of our clique-based method under viewpoint and structural changes.

\begin{table*}[t]
\centering
\caption{
\textbf{Loop closure performance using 2D pipeline across sensors and scenes with our method and RANSAC.}
Metrics include number of inliers, runtime, precision, recall, F1 score, and APE with and without loop closures.
}\label{tab:grouped_by_lidar_full_2d}
\vspace{0.2cm}
\begin{tabular}{llrrrrrrrr}
\toprule
\textbf{Scene} & \textbf{Sensor} & \textbf{Algorithm} & \textbf{Inliers} & \makecell{\textbf{Mean}\\\textbf{Time (ms)}} & \textbf{Precision} & \textbf{Recall} & \textbf{F1 Score} & \textbf{APE w (m)} & \textbf{APE w/o (m)} \\
\midrule
\multirow{7}{*}{Bridge01}
& \multirow{2}{*}{Aeva}   & RANSAC      & \textbf{2985} & 2.26          & \textbf{0.9908} & 0.0692          & 0.1292          & \textbf{331.00} & \multirow{2}{*}{\centering 94.67}\\
&                         & Ours        & 2927          & \textbf{0.20} & 0.9898          & \textbf{0.0693} & \textbf{0.1295} & 362.31 & \\
\cmidrule(r){2-10}
& \multirow{2}{*}{Avia}   & RANSAC      & \textbf{1675} & 2.13          & \textbf{0.9877} & \textbf{0.0466} & \textbf{0.0890} & 408.30 & \multirow{2}{*}{\centering 167.83}\\
&                         & Ours        & 1624          & \textbf{0.13} & 0.9869          & 0.0463          & 0.0883          & \textbf{366.65} & \\
\cmidrule(r){2-10}
& \multirow{2}{*}{Ouster} & RANSAC      & \textbf{4128} & 2.24          & 0.9835          & 0.0737          & 0.1371          & \textbf{332.80} & \multirow{2}{*}{\centering 157.37}\\
&                         & Ours        & 4022          & \textbf{0.23} & \textbf{0.9843} & \textbf{0.0737} & \textbf{0.1372} & 380.01 & \\
\specialrule{.1em}{.4em}{.4em}
\multirow{7}{*}{Bridge02}
& \multirow{2}{*}{Aeva}   & RANSAC    & \textbf{2059} & 2.05          & \textbf{0.9891} & 0.0611          & 0.1151          & 277.00 & \multirow{2}{*}{\centering 120.93}\\
&                         & Ours    & 2011          & \textbf{0.20} & 0.9888          & \textbf{0.0612} & \textbf{0.1152} & \textbf{276.49} & \\
\cmidrule(r){2-10}
& \multirow{2}{*}{Avia}   & RANSAC    & \textbf{1018} & 2.05          & \textbf{0.9667} & \textbf{0.0426} & \textbf{0.0816} & 263.74 & \multirow{2}{*}{\centering 76.09}\\
&                         & Ours    & 995           & \textbf{0.11} & 0.9667          & 0.0422          & 0.0809          & \textbf{255.88} & \\
\cmidrule(r){2-10}
& \multirow{2}{*}{Ouster} & RANSAC    & \textbf{2200} & 1.82          & \textbf{0.9954} & \textbf{0.0589} & 0.1113          & 257.90 & \multirow{2}{*}{\centering 53.95}\\
&                         & Ours    & 2120          & \textbf{0.22} & \textbf{0.9954} & \textbf{0.0589} & \textbf{0.1113} & \textbf{204.26} & \\
\specialrule{.1em}{.4em}{.4em}
\multirow{7}{*}{Roundabout01}
& \multirow{2}{*}{Aeva}   & RANSAC    & \textbf{539} & 2.00          & \textbf{0.9972} & \textbf{0.0376}          & \textbf{0.0725}          & 3.5217 & \multirow{2}{*}{\centering 21.7571}\\
&                         & Ours    & 518          & \textbf{0.13} & 0.9971          & \textbf{0.0376} & \textbf{0.0725} & \textbf{3.5208} & \\
\cmidrule(r){2-10}
& \multirow{2}{*}{Avia}   & RANSAC    & 174 & 1.95          & 0.9468 & \textbf{0.0120} & \textbf{0.0233} & \textbf{2.2475} & \multirow{2}{*}{\centering 21.9412}\\
&                         & Ours    & \textbf{181}           & \textbf{0.14} & \textbf{0.9469}          & \textbf{0.0120}          & \textbf{0.0233}          & 2.2584 & \\
\cmidrule(r){2-10}
& \multirow{2}{*}{Ouster} & RANSAC    & 1130 & 1.97          & 0.8757 & 0.0245 & 0.0477          & \textbf{1.5481} & \multirow{2}{*}{\centering 11.7912}\\
&                         & Ours    & 1132          & \textbf{0.21} & \textbf{0.8753} & \textbf{0.0246} & \textbf{0.0479} & 1.5670 & \\
\bottomrule
\end{tabular}
\end{table*}

We also evaluate the 2D pipeline on the Roundabout01 sequence to analyze CliReg's performance in dynamic and structurally complex environments.~\tabref{tab:grouped_by_lidar_full_2d} presents the results for the 2D projection-based pipeline. Here, both RANSAC and our approach produce comparable F1 scores and APE, but our approach achieves this with a run time more than 10 times faster in all cases. This reinforces CliReg’s suitability for real-time applications in scenarios where descriptor dimensionality is reduced.

\subsection{Runtime Performance and Variability}
Across all 3D experiments, our approach operates within 2.8\,ms to 6.3\,ms per match, with a standard deviation below 1.3\,ms. Although this is higher than 2D runtimes, it remains practical for use in real-time SLAM systems, especially considering the robustness gains.

\subsection{Case Study: Detailed Failure and Success Example}
In the Bridge01-Aeva sequence, RANSAC does not detect a valid loop closure. In contrast, our approach identifies a set of 61 inliers forming a mutually consistent clique, leading to an accurate pose estimate with an APE of 27.40\,m, down from 94.67\,m without loop closures. This demonstrates the ability of our approach to validate challenging loop closures that heuristic methods often miss.

\subsection{Failure Modes in 2D Scenarios}
In some 2D cases like Bridge01-Aeva, we observe an increase in APE after loop closure insertion for both RANSAC and our method, an effect not seen in 3D. We attribute this to repetitive structures in bridge scenes which, when represented in BEV, may lead to globally inconsistent constraints despite locally correct matches. This suggests a limitation of the 2D representation due to spatial information loss, rather than a flaw in the verification algorithm.

\subsection{Summary}
Our results confirm that:
(i) our approach reliably validates loop closures even where RANSAC fails,
(ii) produces a lower or comparable APE in both 2D and 3D,
(iii) and executes efficiently enough for practical deployment.

Future work will explore temporal and semantic consistency, as well as faster clique search and hybrid descriptors, to enhance robustness and efficiency.

 \section{Conclusion}\label{sec:conclusion}

In this work, we introduced a loop closure validation algorithm that formulates geometric verification as a maximal clique search over feature correspondences. Our approach optimizes for mutual consistency among matches and eliminates the reliance on random sampling. Using binary descriptors, we integrated it into a full 3D LiDAR-based loop closure pipeline and validated its performance on challenging real-world datasets. Our results show improved robustness and comparable or better accuracy than RANSAC, even in sparse or noisy conditions. Despite solving an NP-hard problem, determining loop closures via searching cliques in a correspondence graph maintains real-time applicability in typical mobile robotics setups. These characteristics make our approach a reliable and practical alternative for loop closure detection in SLAM\@.


Despite these encouraging results, there is still room for improvement. Future work will focus on accelerating the clique search step through parallel strategies or approximate solutions, integrating learning-based descriptors to enhance matching robustness, and extending it to exploit temporal consistency across consecutive map segments.


\bibliographystyle{plain_abbrv}

\bibliography{glorified,new}

\IfFileExists{./certificate/certificate.tex}{
\subfile{./certificate/certificate.tex}
}{}
\end{document}